\documentclass[conference]{IEEEtran}
\IEEEoverridecommandlockouts
\usepackage{cite}
\usepackage{amsmath,amssymb,amsfonts}
\usepackage{algorithmic}
\usepackage{subfigure}
\usepackage{color}
\usepackage{float}
\usepackage{graphicx}
\usepackage{textcomp}
\usepackage{xcolor}
\usepackage{url}
\def\BibTeX{{\rm B\kern-.05em{\sc i\kern-.025em b}\kern-.08em
    T\kern-.1667em\lower.7ex\hbox{E}\kern-.125emX}}
\begin{document}

\title{Deep Reinforcement Learning \\ for Active High Frequency Trading\\
\thanks{TA and JT acknowledge the EC Horizon 2020 FIN-Tech project for partial support and useful opportunities for discussion. JT acknowledges support from EPSRC (EP/L015129/1). TA acknowledges support from ESRC (ES/K002309/1), EPSRC (EP/P031730/1) and EC (H2020-ICT-2018-2 825215).}
}

\author{
    \IEEEauthorblockN{Antonio Briola\IEEEauthorrefmark{1}, Jeremy Turiel\IEEEauthorrefmark{1}, Riccardo Marcaccioli\IEEEauthorrefmark{2}, Alvaro Cauderan\IEEEauthorrefmark{4}, Tomaso Aste\IEEEauthorrefmark{1}\IEEEauthorrefmark{3}}
    \IEEEauthorblockA{\IEEEauthorrefmark{1}\textit{Department of Computer Science}, \textit{University College London}
    \\ London, UK
    \\\{a.briola, jeremy.turiel.18, t.aste\}@ucl.ac.uk}
    \IEEEauthorblockA{\IEEEauthorrefmark{2}\textit{Chair of Econophysics and Complex Systems}, \textit{Ecole Polytechnique}
    \\ Paris, FR
    \\\{riccardo.marcaccioli\}@ladhyx.polytechnique.fr}
    \IEEEauthorblockA{\IEEEauthorrefmark{3}\textit{Systemic Risk Center}, \textit{London School of Economics}
    \\ London, UK
    \\\{acauderan\}@ethz.ch}
    \IEEEauthorblockA{\IEEEauthorrefmark{4}\textit{Department of Computer Science}, \textit{ETH}
    \\ Zurich, CH}
}

\maketitle

\begin{abstract}
We introduce the first end-to-end Deep Reinforcement Learning (DRL) based framework for active high frequency trading in the stock market. We train DRL agents to trade one unit of Intel Corporation stock by employing the Proximal Policy Optimization algorithm. The training is performed on three contiguous months of high frequency Limit Order Book data, of which the last month constitutes the validation data. In order to maximise the signal to noise ratio in the training data, we compose the latter by only selecting training samples with largest price changes. The test is then carried out on the following month of data. Hyperparameters are tuned using the Sequential Model Based Optimization technique. We consider three different state characterizations, which differ in their LOB-based meta-features.
Analysing the agents' performances on test data, we argue that the agents are able to create a dynamic representation of the underlying environment. They identify occasional regularities present in the data and exploit them to create long-term profitable trading strategies.
Indeed, agents learn trading strategies able to produce stable positive returns in spite of the highly stochastic and non-stationary environment. The source code is available at \url{https://github.com/FinancialComputingUCL/DRL_for_Active_High_Frequency_Trading}.

\end{abstract}


\section{Introduction} \label{intro}

Can machines learn how to trade automatically? What data do they need to achieve this task? If they successfully do so, can we use them to learn something about the price formation mechanism? In this paper, we try to answer such questions by means of a Deep Reinforcement Learning (DRL) approach. 

During the last ten years, DRL algorithms have been applied to a variety of scopes and contexts ranging from robotics \cite{kober2013reinforcement,deisenroth2011pilco} to healthcare \cite{ling2017diagnostic}. For a complete overview of the main application domains we refer the interested reader to the work by Li \cite{li2017deep}. Financial markets are known to be stochastic environments with very low signal to noise ratios, dominated by markedly non-stationary dynamics and characterised by strong feedback loops and non-linear effects \cite{comerton2015dark}. This is true especially for data at highest level of granularity \cite{comerton2015dark, o2015high}, i.e. at what is called the microstructure level. The Limit Order Book (LOB) is the venue where market participants express their intentions to buy or sell a specific amount (volume) of a security at a specific price level by submitting orders at that given price. The interest in applying DRL algorithms to such a context is undeniably strong. First of all, all the algorithms which exploit the power of ANNs are well known to be data hungry models and, thanks to the spreading of electronic trading, everything that happens in the LOB of an assets can be recorded and stored with up to nanosecond resolution. Secondly, several different problems, potentially solvable using Reinforcement Learning, can be defined on such data. 

There are several research questions, with immediate practical interest that must be investigated. Can an agent learn how to assign a given amount of capital among a set of assets? If we need to buy or sell a large amount of assets, can we train an agent to optimally do so? Can we train an agent to directly learn trading strategies at different frequencies and for different numbers of assets?

In the present work, we choose to tackle the latter problem. We show how it is possible to train DRL agents on LOB data to buy and sell single units of a given asset and achieve long term profits. We choose to trade a single unit because our goal is not to maximize the agent's virtual profits, rather to present the first end-to-end framework to successfully deploy DRL algorithms for active High Frequency Trading. This is discussed further in Section \ref{Models}.
Specifically, we apply the Proximal Policy Optimization (PPO) algorithm \cite{schulman2017proximal} to train our agents. We adopt a continual learning paradigm and we tune the hyperparameters of the model through the Sequential Model Based Optimisation (SMBO) \cite{hutter2011sequential} technique. Motivated by the work of Briola \textit{et al.} \cite{briola2020deep}, we use the simplest policy object offered by Stable Baselines \cite{stable-baselines} that implements actor-critic through the usage of a Multilayer Perceptron. Finally, in order to regularize the dynamics of the profit and loss (P\&L) function, we construct the training set by developing a sampling strategy which selects periods of high price activity that are often characterised by higher signal to noise ratios. This provides a more informative training set for the agent.

\section{Background}
\subsection{Limit Order Book}
Nowadays, to facilitate trading, most of the major to minor financial market exchanges employ an electronic trading mechanism called Limit Order Book (LOB). The formation of the price of an asset in a LOB is a self-organising process driven by the submissions and cancellations of orders. Orders can be thought of as visible declarations of a market participant’s intention to buy or sell a specified quantity of an asset at a specified price. Orders are put in a queue until they are either cancelled by their owner or executed against an order of opposite direction. The execution of a pair of orders implies that the owners of such orders trade the agreed quantity of the asset at the agreed price. Order execution follows a \textit{FIFO} (first in, first out) mechanism. An order $x$ is defined by four attributes: the sign or direction $\epsilon_x$ indicates if the owner wants to buy or sell the given asset, the price $p_x$ at which the order is submitted, the number of shares (or volume) $V_x$ that the owner of the order wants to trade and the time $\tau_x$ of submission.
It is therefore defined by a tuple $x = (\epsilon_x, p_x, V_x, \tau_x)$. Whenever a trader submits a buy (respectively, sell) order, an automated trade-matching algorithm checks whether it is possible to execute $x$ against one or more existing orders of the opposite sign with a price smaller or equal to (respectively, greater or equal to) $p_x$. Conventionally, sell orders have a negative signature $\epsilon = -1$ while buy orders have a positive sign $\epsilon = 1$. If any partial matching of the whole volume $V_x$ is possible, it occurs immediately. Instead, any portion of $V_x$ that is not immediately matched becomes active at the price $p_x$, and it remains so until either it is matched to an incoming sell (respectively, buy) order or it is cancelled by the owner.

\begin{figure}[htbp]
\centerline{\includegraphics[scale=0.143]{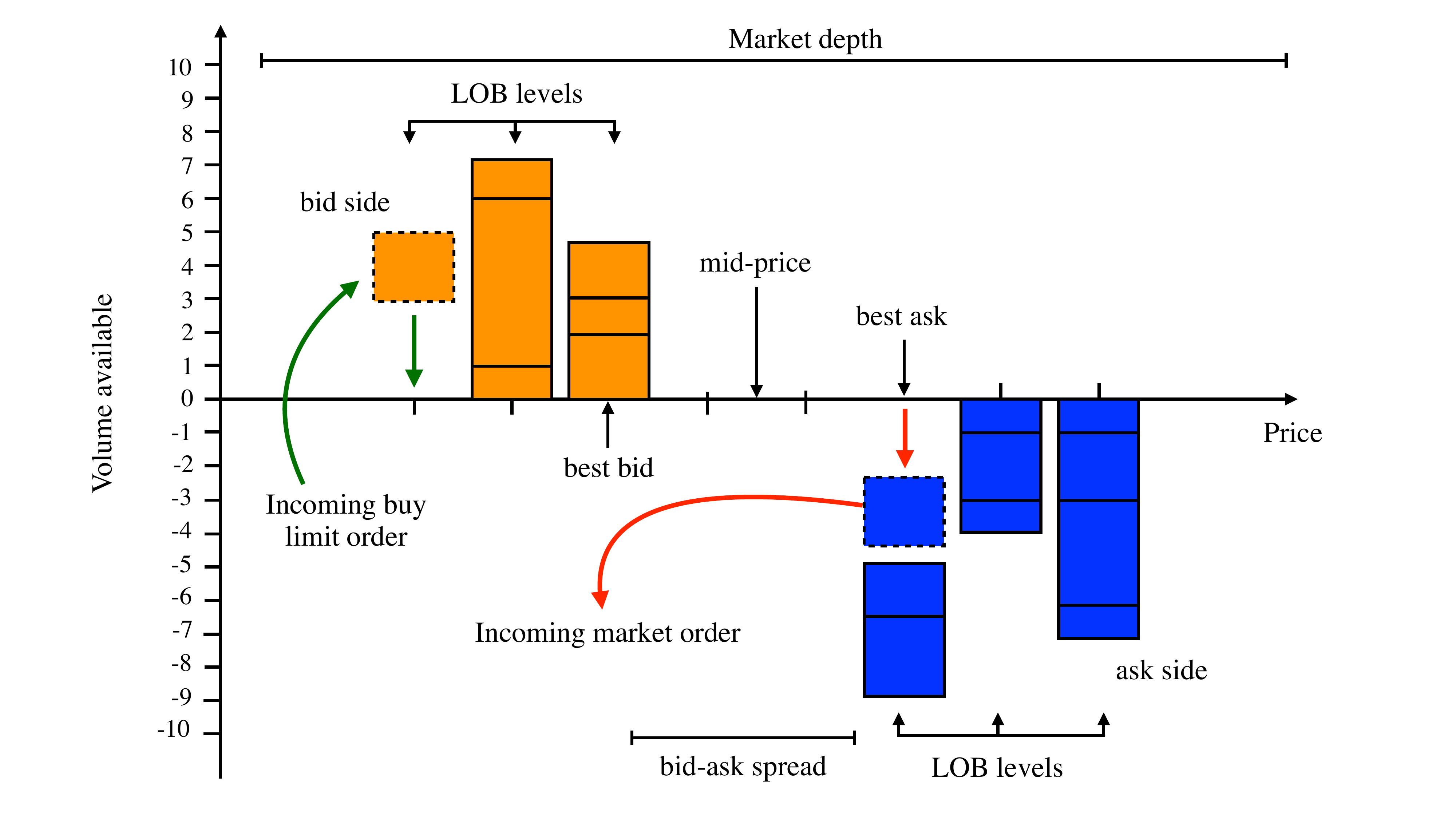}}
\caption{Graphical illustration of a Limit Order Book, its components, related quantities and dynamics.}
\label{fig:LOB_Dynamics}
\vspace{-4mm}
\end{figure}

A schematic representation of a LOB's dynamics is provided in Figure \ref{fig:LOB_Dynamics}. The collection of all active sell limit orders constitutes the ask side of the LOB while the collection of all active buy limit orders forms the bid side. The price difference $\sigma_\tau = p_{a,\tau}^{best} - p_{b,\tau}^{best}$ between the sell and buy orders at their best available price, respectively, (i.e. the sell/buy with lower/greater price) defines the bid-ask spread, while the average between the best bid/ask prices of a security is called the mid-price $m_\tau = (p_{a,\tau}^{best} + p_{b,\tau}^{best})/2$.

Orders that are executed at the best bid/ask price are called market orders (MOs). MOs gain time priority (since their order is executed immediately), but at the cost of paying every unit of the given asset an extra amount at least equal to the spread $\sigma_\tau$. This is the reason why the spread is also said to be the cost of a market order. Orders not entirely executed upon arrival (which therefore sit in full or in part in the LOB until matched or cancelled) are called limit orders. Limit orders sitting at the best bid or ask price are matched with incoming MOs using mechanisms which may differ between markets. The most common priority mechanism (which also rules the dynamics of the LOB considered in this work) is called price-time priority: buy/sell limit orders sitting at the same price are executed by selecting first the limit order with the earliest submission time $\tau_x$. We can therefore define the LOB as the collection of unmatched limit orders for a given asset on a given platform at time $\tau$. The time parameter $\tau$ is naturally a continuous variable which can be studied on its own, e.g. by characterising the distribution of waiting times between two or more types of order or by studying possible seasonalities of one or more LOB observables. However, when someone wants to characterize the effect of a given event on the LOB (like the average price movement after a market order), it is easier to normalize the time and work in tick time. In this system/frame of reference time evolves as a discrete variable which is updated each time an event is recorded on the LOB. For ease of exposition, we terminate here our description of the LOB. However, it can be already noticed how such venue is at the same time a very intriguing yet hard testing ground for RL algorithms. The rules are simple but able to produce complex patterns, the dynamics of the environment are the result of mediated interactions where agents try to cover their real intentions as much as possible and where different agents act according to their own specific strategies and objectives. We refer the interested reader to the book by Bouchaud \textit{et al.} \cite{bouchaud_bonart_donier_gould_2018} for a detailed introduction on the topic and a deep characterization of the dynamics of the various quantities which can be defined on a LOB.

\subsection{Proximal Policy Optimization Algorithm}

Reinforcement Learning (RL) can be described as the problem faced by a learner which attempts to achieve a specific goal while minimizing related costs, through the exploitation of trial-and-error interactions with a dynamic environment. During this learning exercise, the agent experiences different states (i.e. points in the phase space covered by the environment's trajectories) and potentially performs a variety of actions. The effect associated with the specific state-action pair is then evaluated by means of a scoring function and a numerical reward determining the goodness of the chosen action sequence. In this context, the agent needs to find an optimal balance between the exploitation of already experienced state-action pairs and the exploration of unseen ones. Finding this equilibrium is far from trivial. The intrinsic stochasticity involved in solving this maximization problem, makes RL different from other learning paradigms since it does not try to approximate any specific underlying function or maximize a well defined and deterministic score.

This corresponds to finding the optimal mapping (i.e. the policy $\pi$) between a set of states $\mathcal{S}$ and a probability distribution over a set of actions $\mathcal{A}$ able to maximize a discounted, cumulative reward over time $T \in [0, \infty)$. Approaches to solving RL problems can be grouped into different classes. Our focus will be on policy search-based methods, but we refer the interested reader to the book by Richard S. Sutton and Andrew G. Barto \cite{sutton2018reinforcement} for a complete review of the subject.

While effective for a vast range of problems in their basic configuration, RL algorithms have proven to be impacted by the curse of dimensionality \cite{arulkumaran2017deep}. The higher the number of features used to describe the agent's state, the harder is for the agent to learn the optimal strategy. Indeed, it will need to search for it in a space whose size scales combinatorially with the number of features. In order to scale classical RL algorithms to high-dimensional problems, Deep Reinforcement Learning (DRL) was introduced. The usage of Artificial Neural Networks (ANNs) to automatically produce lower-dimensional feature representations from a high dimensional input can allow to overcome the curse of dimensionality. Yet, DRL algorithms can require a large amount of data to properly generalise and the training task is notoriously hard~\cite{li2017deep}.

Among other issues, high sensitivity to hyperparameter tuning and initialization can affect the DRL agent's ability to generalize. For example, a wrong learning rate could push the policy network into a region of the parameter space where it is going to collect the next batch of data under a very poor policy, thereby causing it never to recover. In order to overcome this issue, the Proximal Policy Optimization (PPO) algorithm \cite{schulman2017proximal} has been introduced. PPO is known to be scalable, data efficient, robust \cite{abbeel2016deep} and one of the most suitable for applications to highly stochastic domains. As such, it is a valid candidate for the applicative domain considered here.

The PPO algorithm is part of a family of policy gradients methods for DRL, which are mainly on-policy methods, i.e. they search for the next action based on the present state. In order to do this, policy gradient methods adopt the Policy Loss Function $L^{PG}(Q)$

\begin{equation}
\label{eq:Policy Loss Function}
    L^{PG}(Q)=E_t[\log \pi_Q(a_t|s_t)] \hat{A}_t \; ,    
\end{equation}

where $E_t[\log\pi_Q(a_t|s_t)]$ represents the expected $\log$ probability of taking action $a$ in state $s$ and the estimated Advantage function $\hat A_t$, provides an estimate of the value of the current action compared to the average over all other possible actions at the present state $s$. 

PPO methods ensure the stability of the policy update process by directly controlling the size of the gradient ascent step implied by the Policy Loss Function. In order to do this, the objective function is written as per Equation \ref{eq:Proximal_Policy_Optimization_Loss} 

\begin{equation}
\begin{split}
\label{eq:Proximal_Policy_Optimization_Loss}
    L^{CPI}(Q)= E_t \left[\frac{\pi_Q(a_t|s_t)}{\pi_{Q_{old}}(a_t|s_t)}\hat A_t\right] = E_t[r_t(Q)\hat A_t] \; ,
\end{split}
\end{equation}

where the term $\log\pi_Q(a_t|s_t)$ is replaced by the ratio between the probability of the action under the current policy and the probability of the action under the previous policy $r_t(Q) = \frac{\pi_Q(a_t|s_t)}{\pi_{Q_{old}}(a_t|s_t)}$. This implies that for $r_t(Q) > 1$ the action is more probable in the current policy than in the previous one, while if $0 < r_t(Q) < 1$ the action was more probable in the previous policy than in the current one. We can then notice that when $r_t(Q)$ deviates from $1$ an excessively large policy update would be performed. In order to avoid this, the objective function is modified as per Equation \ref{eq:Proximal_Policy_Optimization_Loss_Clipped} to penalize $r_t(Q)$ values that lie outside of $[1-\epsilon,1+\epsilon]$. We take $\epsilon = 0.2$ here as in the original paper \cite{schulman2017proximal}.

\begin{equation}
\begin{split}
\label{eq:Proximal_Policy_Optimization_Loss_Clipped}
    L^{CLIP}(Q)= E_t \Big[min\Big[\Big(r_t(Q)\hat A_t\Big), \\
    clip \Big(r_t(Q),[1-\epsilon,1+\epsilon]\Big)\hat A_t\Big]\Big] \; .
\end{split}
\end{equation}

The Clipped Surrogate Objective function in Equation \ref{eq:Proximal_Policy_Optimization_Loss_Clipped} takes the minimum between the original $L^{CPI}(Q)$ and the clipped $clip \Big(r_t(Q),[1-\epsilon,1+\epsilon]\Big)\hat A_t$, which removes the incentive for $r_t(Q)$ outside the $[1-\epsilon,1+\epsilon]$ range. The minimum constitutes a lower bound (i.e. pessimistic bound) on the unclipped objective.

\section{Related Work}

As mentioned in Section \ref{intro}, the growth of electronic trading has sparked the interest in DL-based applications to the context of LOB data. The literature on forecasting LOB quantities using supervised learning models is growing rapidly. The work by Kearns and Nevmyvaka \cite{kearns2013machine} presents an overview of AI-based applications to market microstructure data and tasks, including return prediction based on previous LOB states. The first attempt to produce an extensive analysis of DL-based methods for stock price prediction based upon the LOB was recently presented in \cite{tsantekidis2017forecasting}. Starting from a horse-racing type comparison between classical Machine Learning approaches, such as Support Vector Machines, and more structured DL ones, Tsantekidis et al. then considers the possibility to apply CNNs to detect anomalous events in the financial markets, and take profitable positions. The work by Sirignano and Cont \cite{sirignano2019universal} provides a more theoretical approach. It contains a comparison between multiple DL architectures for return forecasting based on order book data, and it discusses the ability of these models to capture and generalise to universal price formation mechanisms throughout assets. More recently, Zhang \textit{et al.} produced two works aimed at forecasting mid-price returns. The first work \cite{zhang2018bdlob} directly forecasts returns at the microstructure level by employing Bayesian Networks, while the second \cite{zhang2019extending} develops a DL architecture aimed at the simultaneous modelling of the return quantiles of both buy and sell positions. The latter introduces the present state-of-the-art modeling architecture combining CNNs and LSTMs to delve deeper into the complex structure of the LOB. Lastly, a recent review of all the aforementioned methods applied to the context of return quantile forecasting can be found in \cite{briola2020deep}, which highlights how modelling the LOB through its temporal and spatial dimensions (with the CNN-LSTM model introduced in \cite{zhang2019extending}) provides a good approximation, but not a necessary nor optimal one. These findings justify the use of an MLP in the present work as the underlying policy network for the DRL agent.

RL applications, involving financial market data, date back to the early 2000s. These typically focused on low frequency single asset (or index) trading \cite{moody2001learning, moody1999reinforcement}, portfolio allocation \cite{moody1998performance, neuneier1996optimal, mihatsch2002risk}, up to early forms of intra-day trading \cite{moody1999minimizing}. For a complete review of the subject, we refer the interested reader to \cite{meng2019reinforcement}. The recent spike of interest in DRL and its various applications has naturally brought researchers to apply such algorithms to financial market data or simulations \cite{byrd2019abides, karpe2020multi}. Concerning trading applications, researchers mostly looked at active trading on single assets with simulated or real data in the low frequency (typically daily) domain \cite{bertoluzzo2012testing, chen2019application, dang2019reinforcement, deng2016deep, jeong2019improving, kim2017intelligent, zhang2020deeptrading, theate2020application}. Other works focused on a variety of tasks such as multi-asset allocation \cite{yang2020deep, jiang2017cryptocurrency, zhang2020foliodeep} or portfolio management in general \cite{hu2019deep}. The body of literature involving DRL applications to the high-frequency domain is surprisingly scarce. Besides recent works aiming to apply DRL algorithms to market making on simulated high frequency data \cite{lim2018reinforcement, wang2019electronic}, no literature on DRL applications to real high-frequency LOB data is present. In the present work we try to fill this gap and propose the first end-to-end framework to train and deploy a Deep Reinforcement Learning agent on real high frequency market data for the task of active high frequency trading.

\section{Methods}

\subsection{Data}
\label{Data}
High quality LOB data from the LOBSTER dataset \cite{lobster_data} represent the cornerstone of the research presented in the current work. This is a widely adopted dataset in the market microstructure literature \cite{balch2019evaluate, bibinger2019estimation, bouchaud_bonart_donier_gould_2018, bibinger2016volatility} which provides a highly detailed, event-by-event description of all micro-scale market activities for each stock listed on the NASDAQ exchange. The dataset lists every market order arrival, limit order arrival and cancellation that occurs on the NASDAQ platform between 09:30am – 04:00pm on each trading day. Trading does not occur on weekends or public holidays, so these days are excluded from all the analyses performed in the present work.  A minimum tick size of $\theta=0.01\$$ is present for all stocks in the NASDAQ exchange. The flow of orders, together with their size and side, is recorded in the event files. LOBSTER also provides the derived LOB aggregation, which forms the input data for the present work. For each active trading day, the evolution of the LOB up to 10 levels is provided. It is worth noting that, each order's dollar price is multiplied by 10000. Experiments described in the next sections are performed using the reconstructed LOB of Intel Corporation's stock (INTC), which is known to be one of the most traded large tick stocks on the NASDAQ exchange \cite{bouchaud_bonart_donier_gould_2018}. While high market activity may aid DRL training (the agents will explore a wider region of the phase space) or not (lower signal to noise ratio as the price process is more efficient), the fact that we have chosen to consider a large tick stock is definitively beneficial. Indeed, prices of large tick stocks, i.e. stocks with wide price levels, are known to be less stochastic and more dependent on the underlying available LOB information than small tick stocks \cite{bouchaud_bonart_donier_gould_2018}. Further, as each tick is of meaningful size, the LOB is dense which allows to pass volumes alone to the DRL agent without loss of information.

The whole training dataset consists of 60 files (one per trading day between 04-02-2019 and 30-04-2019). The validation dataset consists of 22 (one per trading day between 01-05-2019 and 31-05-2019). The test dataset consists of 20 files (one per trading day between 03-06-2019 and 28-06-2019). All the experiments presented in the current work are conducted on snapshots of the LOB with a depth (number of tick size-separated limit order levels per side of the Order Book) of $10$. Due to the widely different dynamics and higher volatility during the market open and market close periods, these are excluded from both the training and test set for each day (first and last $2 \times 10^5$ ticks). This is done in order to facilitate the agent's learning under \textit{normal} market conditions.

\subsection{Models}
\label{Models}
In the present work, we consider three different training and testing scenarios $c_{i \in [201, 202, 203]}$,  which only differ in the way the DRL agents' state $\mathcal{S}$ is characterized:

\begin{itemize}
\item $\mathcal{S}_{c_{201}}$: The state of the agent at time $\tau$ is defined by the LOB's volumes for the first ten levels on both the bid and ask sides. In addition, the agent is provided the LOB states for the last 9 ticks and the agent's current position (long, short, neutral). 
\item $\mathcal{S}_{c_{202}}$: The state of the agent at time $\tau$ is defined as per $\mathcal{S}_{c_{201}}$, with the addition of the mark to market value for the agent's current open position (zero if not in a position, by definition). The mark to market value is defined as the profit that the agent would obtain if it decides to close its position at the current time. This allows the agent to evaluate an action with knowledge of how its trade has been performing and what closing the position might imply in terms of reward and, hence, return. Moreover, an improved return profile with respect to $\mathcal{S}_{c_{201}}$ would suggest that the optimal exit of a trade is not independent of its return at the present time.
\item $\mathcal{S}_{c_{203}}$: The state of the agent at time $\tau$ is defined as per $\mathcal{S}_{c_{202}}$, with the addition of the current bid-ask spread. This allows the agent to evaluate the trading costs before opening/closing a position. These impact the profitability of the potential trade, and hence the ability to estimate the potential net return of a trade.
\end{itemize}

For all the different state definitions, at each tick time $\tau$, the agent must choose to perform a specific action $ a \in \mathcal{A}$. The set $\mathcal{A}$ of all actions available to the agent is $\mathcal{A} = \{a_i\}_{i=0}^3 = \{ \text{\textit{sell}}, \, \text{\textit{stay}}, \, \text{\textit{buy}}, \, \text{\textit{daily\_stop\_loss}}\}$. The outcome of each action depends on the agent's position $\mathcal{P} = \{ \mathfrak{p}_{i}\}_{i \in{N, L, S}} = \{\text{\textit{neutral}}, \, \text{\textit{long}}, \, \text{\textit{short}} \, \}$.  A position-action pair $(\mathfrak{p}_i, a_i)$ fully defines how the agent will interact with the LOB environment. We hence list the possible pairs and their respective outcomes.

\begin{itemize}
\item $(\mathfrak{p}_N, \, a_0)$: The agent decides to sell (i.e. $a_0$), even if it does not currently own the stock (i.e. $N$). As such, it decides to enter a short position $\mathfrak{p}: N \to S$ for one unit of the underlying stock. Entering a short position is the decision to sell the stock at the current best bid price to buy it back at a future time for the market price available at that time. This action is based on the expectation that the price will fall in the future.

\item $(\mathfrak{p}_N, \, a_2)$: The agent decides to buy (i.e. $a_2$), while not owning the stock at present (i.e. $N$). As such, it decides to enter a long position $\mathfrak{p}: N \to L$ for one unit of the underlying stock. Entering a long position is the decision to buy the stock at the current best ask price to sell it at a future time for the market price available at that time. This action will probably be based on the expectation that the price will rise in the future.

\item $(\cdot, a_1)$: Independently on its state, if the agent decides to sit $A_1$, it will not perform any action.

\item $(\mathfrak{p}_S, a_0)$: The agent decides to sell (i.e. $a_0$), even if it has already a short position open (i.e. $S$). The position is kept open and no additional position is taken on (no change). This sparks from the essential approach of this work, where the agent can only take unit-stock positions.

\item $(\mathfrak{p}_L, a_2)$: The agent decides to buy (i.e. $a_2$), even if it has already a long position open (i.e. $L$). The position is kept open and no additional position is taken on (no change). This sparks from the essential approach of this work, where the agent can only take unit-stock positions.

\item $(\mathfrak{p}_L, a_0)$: The agent decides to sell (i.e. $a_0$) while in a long position (i.e. $L$). The position is closed and restored to null $\mathfrak{p}: L \to N$. The profit associated with the trade is hence calculated and given to the agent as feedback for its action.

\item $(\mathfrak{p}_S, a_2)$: The agent decides to buy (i.e. $a_2$) while in a short position (i.e. $S$). The position is closed and restored to null $\mathfrak{p}: S \to N$. The profit associated with the trade is hence calculated and given to the agent as feedback for its action.

\item $(\cdot, a_3)$: The agent checks the current cumulative profit for the day $R_{day}$; if $R_{day} < 0$ the current position is closed (daily stop-loss action $a_3$) and further trading for the day is disabled to avoid excessive losses, otherwise a sit action (i.e. $a_1$) is performed.

\end{itemize}

As described above, for some position-action pairs a non-zero profit (or loss) is experienced by the agent. The profit associated with closing at time $\tau$ a long/short position opened at time $\tau-t$ is intuitively defined as $R_{l/s} =  p^{best}_{a/b,\tau} - p^{best}_{b/a,\tau-t}$, i.e. the net dollar return of the trade (price difference minus cost of trade). Incorporating the spread crossing dynamics (trading at bid/ask rather than mid price) directly incorporates transaction costs and makes the experiment more realistic.

Further, the rationale for introducing the \textit{daily\_stop\_loss} action is that the agent may decide to skip an entire environment (day) if, given the current policy, it is too hard to implement a profitable trading strategy.


The reward function $r_{e,E,d}$ (experienced by the agent at the end of the $e$-th tick of the $E$-th epoch of the $d$-th trading day) is a function of the action-state pair which coincides with the profit $R_{l/c}$ for all actions except the stop-loss.

At this point a final observation is due. The careful reader will have noticed that the agent we define is able to trade and hold in inventory only one unit of the underlying asset at each point in time. Even though this may seem like an unrealistic assumption which affects the applicability of our method, we argue just the opposite.

Every action that an agent performs on a LOB will alter the LOB itself and influence the future actions and reactions of other market participants. Every time we buy (sell) one or more units of an asset, we are consuming liquidity at the ask (bid). This simple action creates a surplus of demand which is going to push the price up (down) proportionally to the square root of the size of the order~\cite{bouchaud_bonart_donier_gould_2018}.
By trading only one unit we are sure that there is enough liquidity (as the minimum is one unit) for the agent to execute at the given best bid/ask price without the need to consume the liquidity at higher levels in the LOB (which would worsen the execution price and impact returns). Further, larger trades are usually broken down into smaller units which are executed separately. A variable order size would imply the need for a more complex execution strategy and the requirement to account for self-impact on the price of execution as the orders are placed by the agent. Hence, our choice of single unit trading supports the simplicity and applicability of our framework and its results to live scenarios.

\subsection{Training-Test Pipeline}

All experiments are organised into three separate stages.

\begin{enumerate}
    
    \item The training phase is performed using a continual learning approach. For each one of the 82 available trading days, 5 LOB snapshots of $10^4$ consecutive ticks each are selected. The snapshots include the largest absolute differences between mid-prices in the day. Among them, only 25 snapshots are sampled and transformed into vectorised DRL environments \cite{stable-baselines}. This sampling procedure allows to choose data containing high signal-to-noise ratios and price changes which are more likely to be actionable. Each environment is then run over $30$ epochs. The underlying agent's policy network is represented by a simple Multilayer Perceptron which consists of two hidden layers of 64 neurons each, shared between the action network and the value network. The last hidden layer is then mapped directly to the two final single-valued network outputs (action and value).
    
    \item A Bayesian Optimisation (BO) \cite{shahriari2015taking, frazier2018bayesian, archetti2019bayesian, candelieri2019global} of specific model hyperparameters (i.e. the learning rate and the entropy coefficient) is performed on the validation set whose environments (i.e. LOB snapshots) are sampled analogously to the training ones. A Gaussian Process (GP) regressor with a Squared Exponential kernel \cite{rasmussen2003gaussian} is used to fit the unknown function which maps hyperparameter values to out of sample rewards, Based on the cumulative profit registered over all the validation trading days, the next realization to be evaluated is iteratively chosen exploiting the Expected Improvement function \cite{movckus1975bayesian, jones1998efficient}. The fitted function then yields the optimal hyperparameter values for training.

    \item The policy learned by the agent is independently tested on all days of the test set during the last stage. The cumulative daily profit $R_{\text{day}}$ is updated each time a position is closed and the corresponding reward is recorded in order to define a P\&L trajectory. Obtained results are hence used to compare the effect of the different state characterizations on the agent's out of sample performance.

\end{enumerate}

\section{Results and Discussion}
We report out of sample (test) results for ensembles of DRL models defined according to the state definitions in Section \ref{Models}. To compare performances we consider the following perspectives on the returns generated by the agent: cumulative P\&L across test days, daily P\&L mean and standard deviation across different iterations and return distribution for the agents' trades across test days and iterations.

\begin{figure*}[ht]
    \centering
    \subfigure[]
    {
        \includegraphics[width=0.3\linewidth]{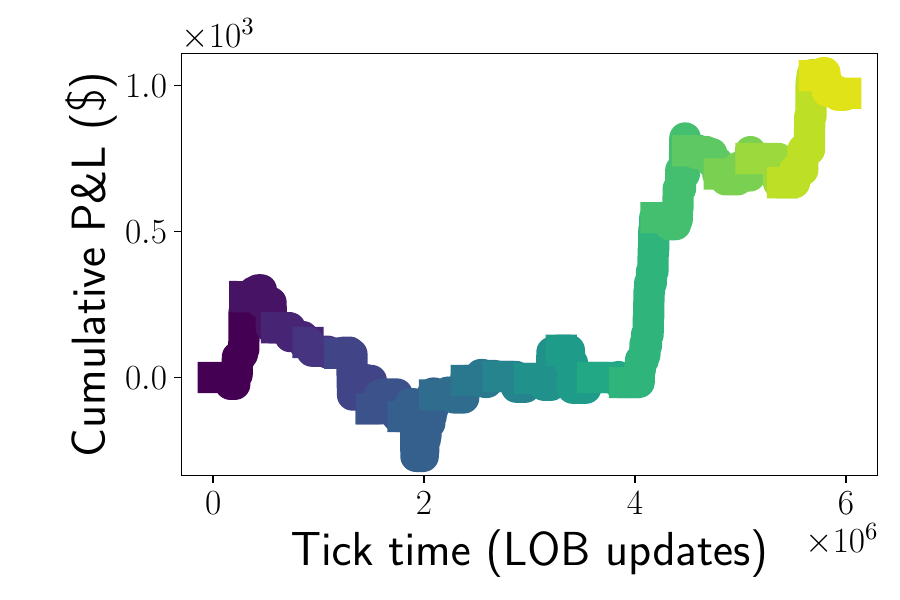}
        \label{fig:only_pos_cum}
    }
    \subfigure[]
    {
        \includegraphics[width=0.3\linewidth]{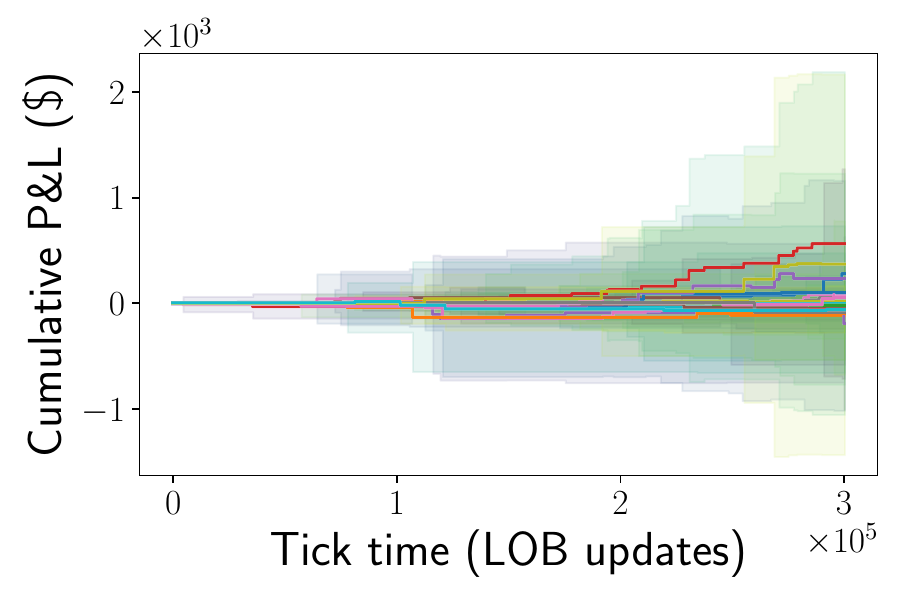}
        \label{fig:only_pos_day}
    }
    \subfigure[]
    {
        \includegraphics[width=0.3\linewidth]{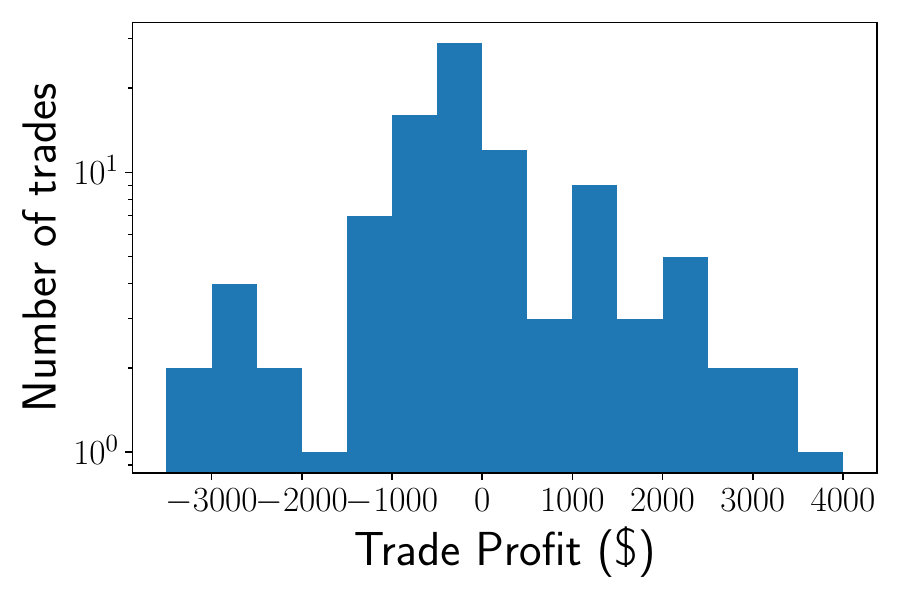}
        \label{fig:only_pos_hist}
    }
    \caption{Cumulative mean return across the 30 ensemble elements (a), daily mean cumulative return and standard deviation (b) and trade return distribution (c) for the full test (20 days) and agent state definition $\mathcal{S}_{c_{201}}$. Notice that the values of P\&L directly depend on the fact that, in LOBSTER data, each order’s dollar price is multiplied by 10000 (see Section \ref{Data}).}
    \label{fig:only_pos}
    \vspace{-2mm}
\end{figure*}

\begin{figure*}[!ht]
    \centering
    \subfigure[]
    {
        \includegraphics[width=0.3\linewidth]{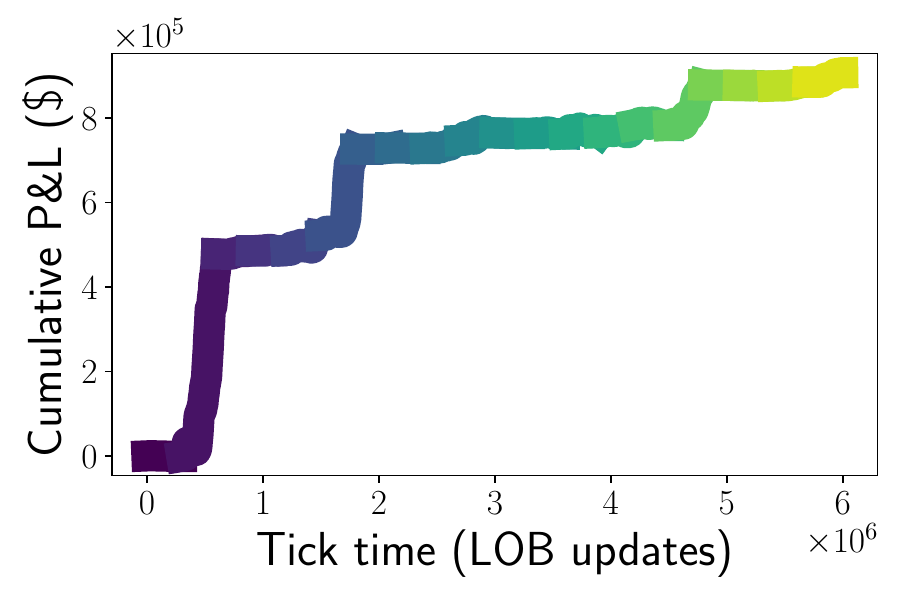}
        \label{fig:pos_mtm_cum}
    }
    \subfigure[]
    {
        \includegraphics[width=0.3\linewidth]{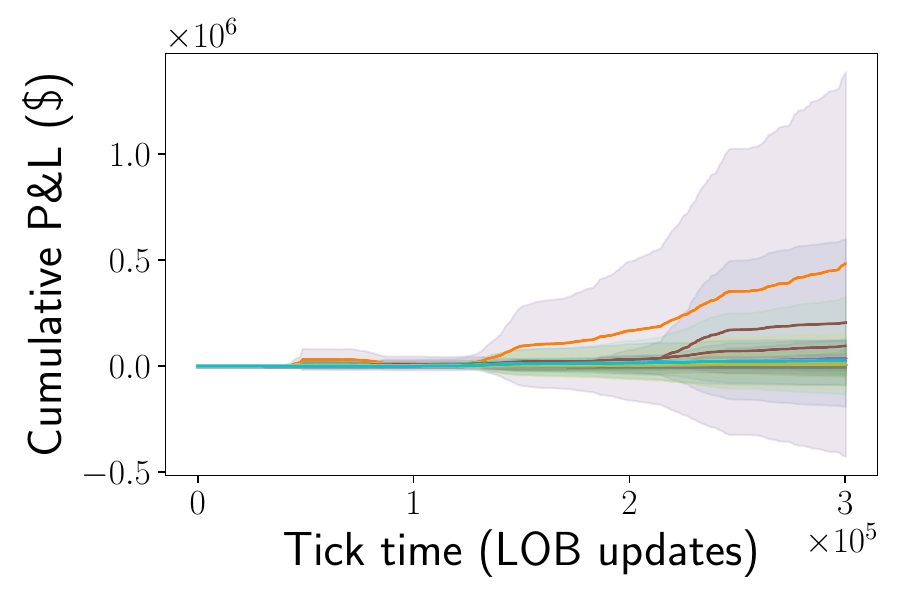}
        \label{fig:pos_mtm_day}
    }
    \subfigure[]
    {
        \includegraphics[width=0.3\linewidth]{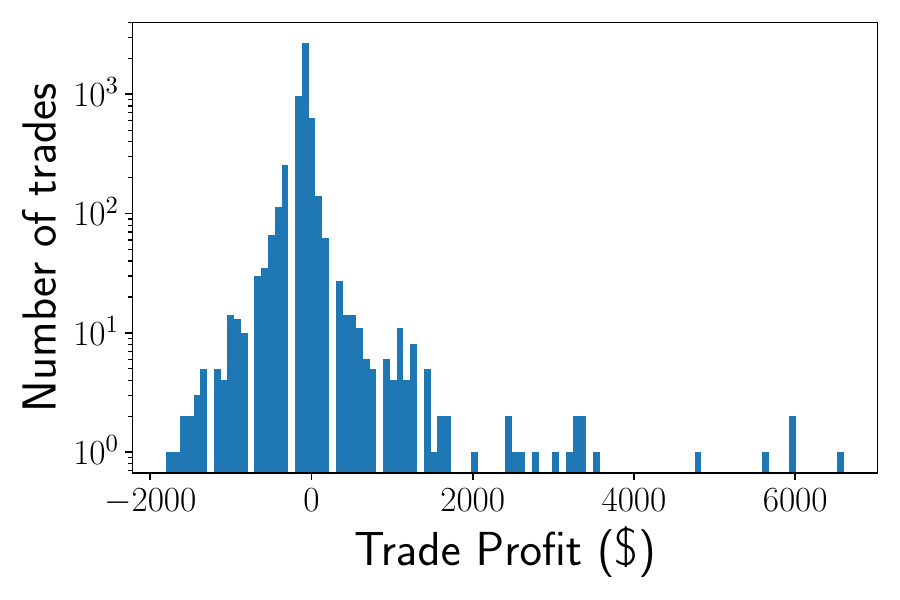}
        \label{fig:pos_mtm_hist}
    }
    \caption{Cumulative mean return across the 30 ensemble elements (a), daily mean cumulative return and standard deviation (b) and trade return distribution (c) for the full test (20 days) and agent state definition $\mathcal{S}_{c_{202}}$. Notice that the values of P\&L directly depend on the fact that, in LOBSTER data, each order’s dollar price is multiplied by 10000 (see Section \ref{Data}).}
    \label{fig:pos_mtm}
    \vspace{-2mm}
\end{figure*}

\begin{figure*}[!ht]
    \centering
    \subfigure[]
    {
        \includegraphics[width=0.3\linewidth]{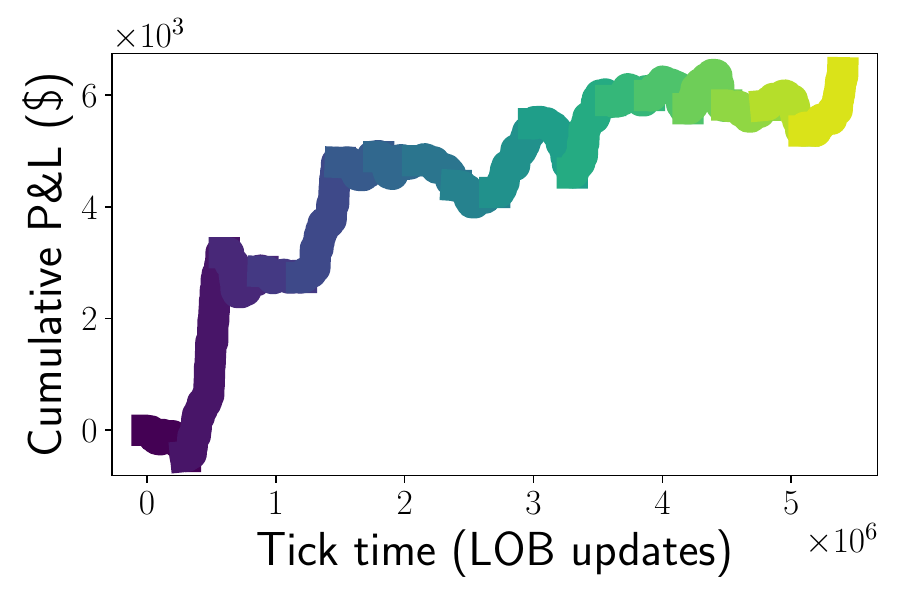}
        \label{fig:pos_mtm_spread_cum}
    }
    \subfigure[]
    {
        \includegraphics[width=0.3\linewidth]{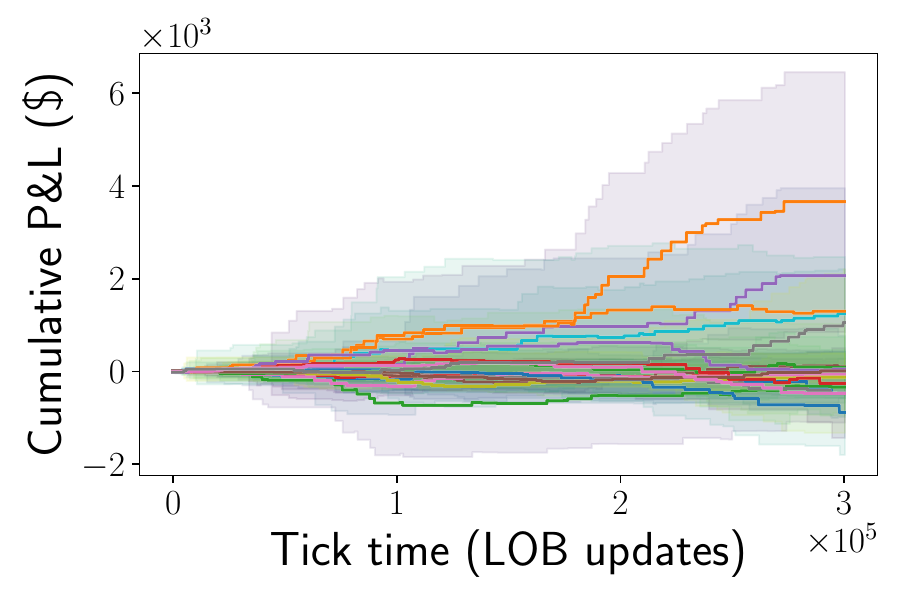}
        \label{fig:pos_mtm_spread_day}
    }
    \subfigure[]
    {
        \includegraphics[width=0.3\linewidth]{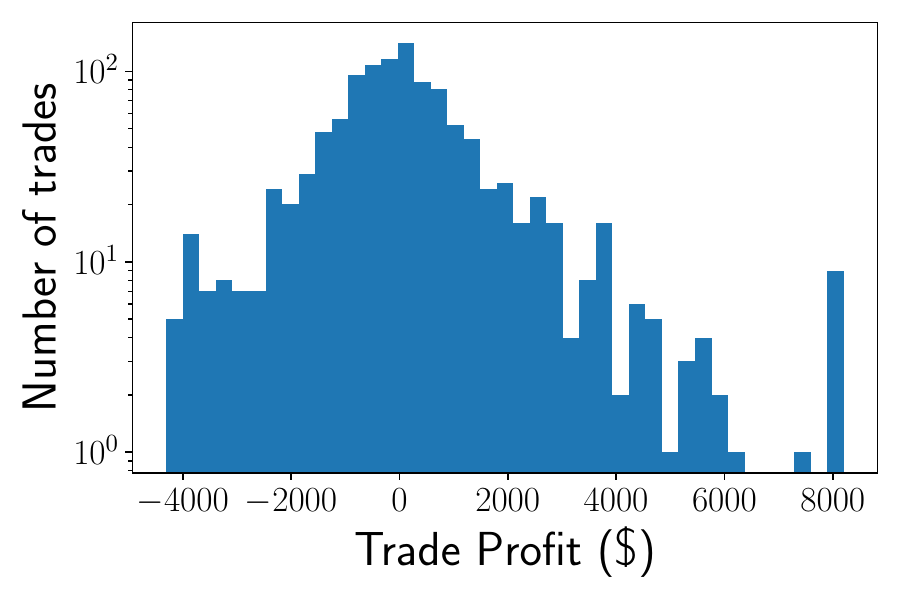}
        \label{fig:pos_mtm_spread_hist}
    }
    \caption{Cumulative mean return across the 30 ensemble elements (a), daily mean cumulative return and standard deviation (b) and trade return distribution (c) for the full test (20 days) and agent state definition $\mathcal{S}_{c_{203}}$. Notice that the values of P\&L directly depend on the fact that, in LOBSTER data, each order’s dollar price is multiplied by 10000 (see Section \ref{Data}).}
    \label{fig:pos_mtm_spread}
    \vspace{-2mm}
\end{figure*}

Figures \ref{fig:only_pos}, \ref{fig:pos_mtm}, \ref{fig:pos_mtm_spread} present results for the three analyses described above and the three state characterizations from Section \ref{Models}.

We start by noticing that the agents are able to produce profitable strategies (net of trading costs) on average across their ensembles for each state characterization. From Figures \ref{fig:only_pos_cum}, \ref{fig:pos_mtm_cum}, \ref{fig:pos_mtm_spread_cum} we observe a significant improvement both in magnitude of the cumulative profit as well as in the more persistent upward trend in time between $\mathcal{S}_{c_{201}}$ and $\mathcal{S}_{c_{202}}$. $\mathcal{S}_{c_{203}}$ shows good performance when compared to $\mathcal{S}_{c_{201}}$, while underperforming $\mathcal{S}_{c_{202}}$. This suggests that the mark to market added to the $\mathcal{S}_{c_{202}}$ state definition is highly informative of the agent's expected reward and strongly improves convergence and performance. On the other hand the spread, which is included in $\mathcal{S}_{c_{203}}$, does not improve performance, likely due to the fact that this is mostly constant in large tick stocks. The outperformance of $\mathcal{S}_{c_{203}}$ when compared to $\mathcal{S}_{c_{201}}$ is likely due to the state definition including the mark to market.
Future work should investigate a larger sample of environment definitions, as this is beyond the scope of this more foundational and elementary work.

The plots in Figures \ref{fig:only_pos_day}, \ref{fig:pos_mtm_day}, \ref{fig:pos_mtm_spread_day} show cumulative daily mean profits with confidence intervals of one standard deviation across ensemble elements per tick. We observe a positive skew of the means and of the confidence intervals as well as a trend in time, with specific days being more volatile than others. These plots allow to see the agents' performances across different environments (i.e. market conditions) and clearly show that profits are not characterised by daily seasonalities or concentrated in very few days alone. Indeed we notice throughout that most of the days in the test set are characterised by positive returns and that positive returns are larger than negative ones, on average.

We observe that $\mathcal{S}_{c_{202}}$ reaches P\&L levels an order of magnitude above the other state characterizations. This result suggests that more trading opportunities are exploited by the agent. This can be understood as the agent having information about its current unrealised profit which allows it to trade in and out of positions more smoothly. Indeed, we observe smoother curves in the daily return plots and a number of trades two orders of magnitude higher than $\mathcal{S}_{c_{201}}$ (and one order of magnitude higher than $\mathcal{S}_{c_{203}}$).

When looking at the return distributions for each trade performed by the agent across days and ensemble realisations (histograms in Figures \ref{fig:only_pos_hist}, \ref{fig:pos_mtm_hist}, \ref{fig:pos_mtm_spread_hist}) we immediately notice the different levels of trading frequencies across state characterizations. As discussed above, information about the current unrealised profit allows the agent to manage its trading activity more accurately and frequently. Yet, the realised profit distributions are broadly similar in terms of the shape of the body and the presence of tails with positive skewness and a peak around zero. The cutoff on the negative distribution tail may be attributed to the day stop loss action available to the agent. As expected, the nearly symmetric distribution body observable in Figures \ref{fig:pos_mtm_hist}, \ref{fig:pos_mtm_spread_hist} suggests a limited price predictability. Indeed, many trades are abandoned early with equally likely positive and negative outcomes. The likely difference arises for larger returns for which the authors suggest the following possible explanations: the spread (cost of trading) has less impact, larger price moves arise from inefficiencies which are exploitable and that, in the absence of meaningful price predictability, good position management allows to obtain positive P\&L. The latter is likely applicable to $\mathcal{S}_{c_{202}}$, $\mathcal{S}_{c_{203}}$, while rare inefficiencies might be the source of profit for $\mathcal{S}_{c_{201}}$. Yet, this directly shows how markets are inefficient (not in the mere sense of price predictability, but rather of being exploitable for profit) at least for short periods of time in the microstructural domain. This finding is likely due to the strategy being based on innovative DRL techniques which are not (yet) widely implemented in this domain and hence have not exploited and removed these inefficiencies.

When looking at the body of the distributions and their tails, there is no strong evidence to justify the widely different return profiles between different trading state characterizations. What seems to really drive the different return profiles, is the number of trades, as reflected in the histogram plots. Indeed, agents with knowledge of their mark to market have exact information about their potential reward and only need to evaluate whether the potential for further upside is worth keeping the position open. This allows to exploit more trading opportunities which lead to an increase in P\&L with no apparent effect on the quality and profitability of individual trades. Indeed this suggests a characteristic trade and return horizon throughout strategies, perhaps characteristic of the asset and its price dynamics.

\section{Limitations}
The method presented in the current research work presents some limitations. One of them is its sample inefficiency, a common issue in RL. The fact that the agent receives a reward only when closing a position means that feedback is only as frequent as the agent's trade. In our case, positions are held for tens, hundreds, or even thousands of time steps, hindering convergence speed and leading to a sparse reward function. Furthermore, considering the task's challenging nature and the reward's sparseness, we observe that, sporadically, the agent needs to properly initialize (i.e. it converges to sub-optimal policy such as not trading at all). This finding is expected and results from the PPO algorithm's exploration policy and the negative expected return the agent experiences at the beginning of the exploration stage. A few solutions could alleviate these issues, such as temporal aggregation of the input time series to reduce the sparseness of the reward or the usage of different RL algorithms with a better-suited exploration policy, but we leave this for future work.

\section{Conclusions}

In the present work we showed how to to successfully train and deploy DRL models in the context of high frequency trading. We investigate how three different state definitions affect the out-of-sample performance of the agents and we find that the knowledge of the mark-to-market return of their current position is highly beneficial both for the global P\&L function and for the single trade reward. However, it should be noticed that, independently on the state definition, the agents are always able to \textquotedblleft beat the market\textquotedblright and achieve a net positive profit in the whole training sample considered and in most of the single trading days it is composed of.

Even though we have considered an agent who buys at the best ask and sells at the best bid (and therefore needs to pay the spread to close a position), the exercise here performed is only a first attempt to realistically employ DRL algorithm in a markedly non-stationary environment as the Limit Order Book and financial markets. First of all, we considered an agent who is able to trade only one single unit of the underlying asset. As a result, we safely considered zero the impact of its own trades on the market. However, it is well known that in a realistic setting, where multiple units of the underlying asset can be bought and sold, the impact of our own trades influences in a severe way the final profit of a strategy. 
Nevertheless, the very own fact that, even in a simplistic setting, a DRL agent can potentially perform well in a highly stochastic and non-stationary environment is worth noticing per se. Further, the ability to produce out of sample profitable strategies shows that temporary market inefficiencies exists and can be exploit to produce long lasting profits and therefore adds a proof against the long-lasting belief that markets are intrinsically efficient.

\bibliographystyle{IEEEtran}
\bibliography{my_bib.bib}
\end{document}